\documentclass{article} 

\usepackage{url}
\usepackage{breakurl}
\usepackage[breaklinks]{hyperref}

\usepackage{comment}
\usepackage{amssymb,amsmath,amsfonts,bm}
\usepackage{graphicx}
\usepackage{caption}
\usepackage{booktabs}
\usepackage{color}
\usepackage{booktabs} 
\usepackage{subfigure}
\usepackage{comment}
\usepackage{makecell}

\usepackage[title]{appendix}

\usepackage{hyperref}

\usepackage{amsmath,amsfonts,bm}

\def\eqref#1{equation~\ref{#1}}

\def\1{\bm{1}}

\DeclareMathAlphabet{\mathsfit}{\encodingdefault}{\sfdefault}{m}{sl}
\SetMathAlphabet{\mathsfit}{bold}{\encodingdefault}{\sfdefault}{bx}{n}

\def\gL{{\mathcal{L}}}

\def\gN{{\mathcal{N}}}

\newcommand{\E}{\mathbb{E}}

\newcommand{\KL}{D_{\mathrm{KL}}}

\newcommand{\eat}[1]{}

\usepackage[accepted]{icml2019}

\icmltitlerunning{Latent Translation: Crossing Modalities by Bridging Generative Models}

\begin{document}

\twocolumn[
\icmltitle{Latent Translation: \\ Crossing Modalities by Bridging Generative Models}

\begin{icmlauthorlist}
\icmlauthor{Yingtao Tian}{stonybrook}
\icmlauthor{Jesse Engel}{googleai}
\end{icmlauthorlist}

\icmlaffiliation{stonybrook}{Department of Computer Science, Stony Brook University, Stony Brook, NY, USA}
\icmlaffiliation{googleai}{Google AI, San Francisco, CA, USA}

\icmlcorrespondingauthor{Yingtao Tian}{yittian@cs.stonybrook.edu}
\icmlcorrespondingauthor{Jesse Engel}{jesseengel@google.com}

\icmlkeywords{Machine Learning, ICML}  

\vskip 0.3in
]

\printAffiliationsAndNotice{}

\begin{abstract}
End-to-end optimization has achieved state-of-the-art performance on many specific problems, but there is no straight-forward way to combine pretrained models for new problems.
Here, we explore improving modularity by learning a post-hoc interface between two existing models to solve a new task.
Specifically, we take inspiration from neural machine translation, and cast the challenging problem of cross-modal domain transfer as unsupervised translation between the latent spaces of pretrained deep generative models.
By abstracting away the data representation, we demonstrate that it is possible to transfer across different modalities (e.g., image-to-audio) and even different types of generative models (e.g., VAE-to-GAN).
We compare to state-of-the-art techniques and find that a straight-forward variational autoencoder is able to best bridge the two generative models through learning a shared latent space.
We can further impose supervised alignment of attributes in both domains with a classifier in the shared latent space. 
Through qualitative and quantitative evaluations, we demonstrate that locality and semantic alignment are preserved through the transfer process, as indicated by high transfer accuracies and smooth interpolations within a class.
Finally, we show this modular structure speeds up training of new interface models by several orders of magnitude by decoupling it from expensive retraining of base generative models. 
\end{abstract}

\section{Introduction} \label{sec:introduction}

\begin{figure}[ht]
\centering
\includegraphics[width=0.95\columnwidth,trim={0 85pt 360pt 0},clip]{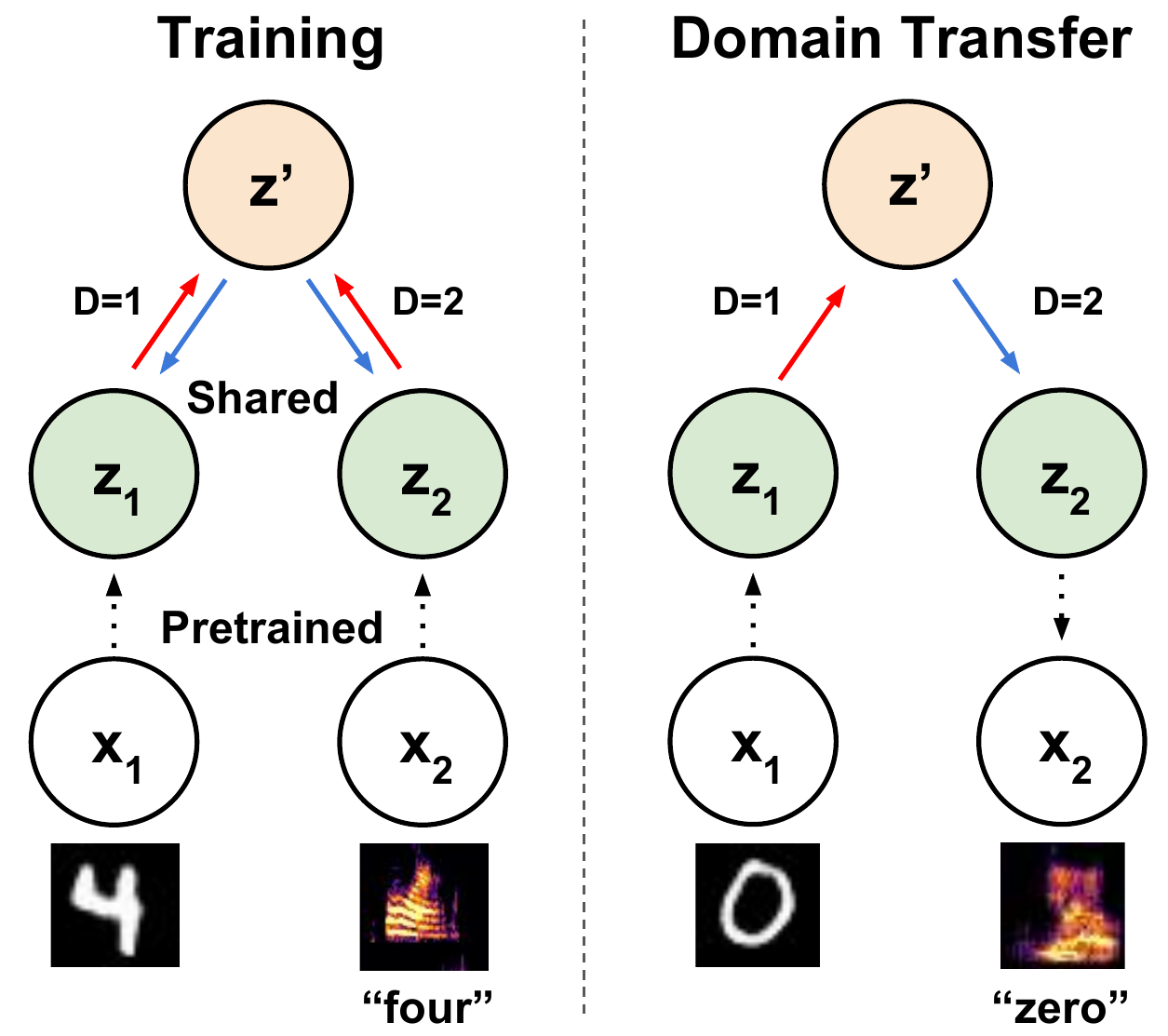}
\caption{
Latent translation with a shared autoencoder. Pretrained generative models provide embeddings ($z_1$, $z_2$) for data in two different domains ($x_1$, $x_2$), 
here shown as written digits and (spectrograms of) spoken digits. 
A shared autoencoder creates joint embeddings ($z'$) which are encouraged to overlap by a sliced-wasserstein distance and semantically structured by a linear classifier. The autoencoder is trained with an additional one-hot domain label ($D$), and domain transfer occurs by encoding with one domain label and decoding with the other.
More details are available in Section~\ref{sec:method}.
}
\label{fig:overall}
\end{figure}

\begin{figure*}[t]
\centering
\includegraphics[width=0.9\textwidth,trim={0 140pt 0 10pt},clip]{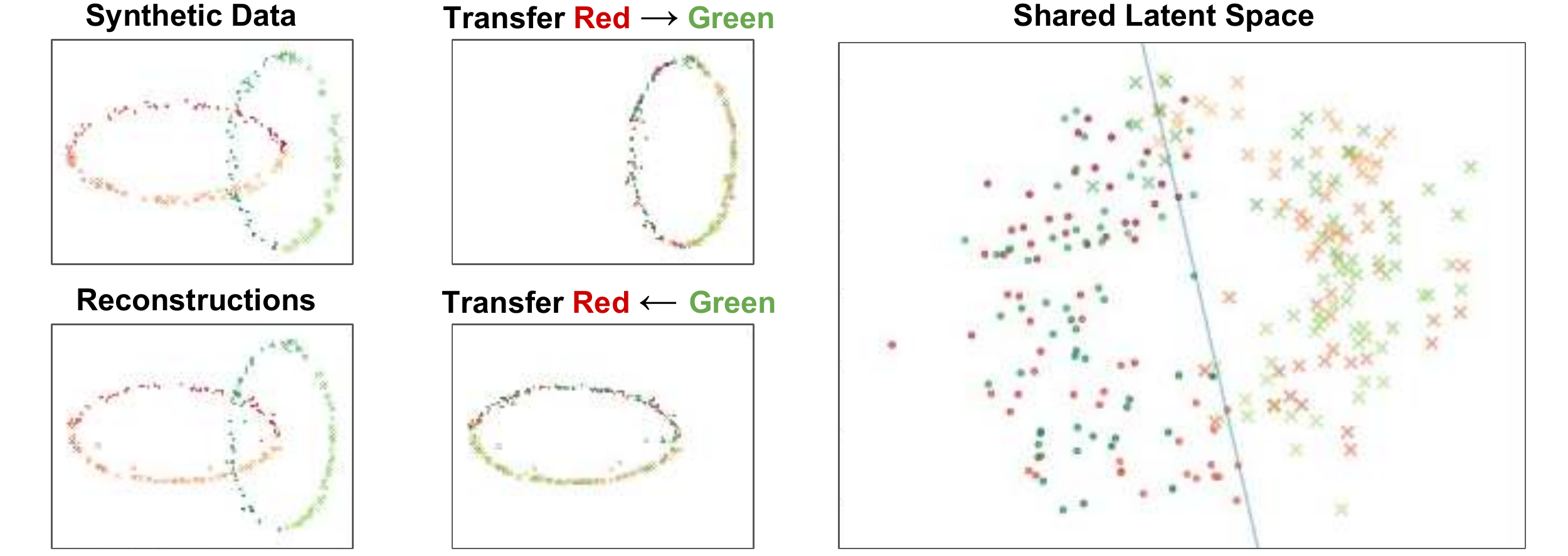}
\caption{
Synthetic problem demonstrating latent translation (best viewed in color). 
Synthetic datasets are created to represent pretrained embeddings for two data domains (red and green ellipses, 2-dimensional). 
A small "bridging" autoencoder (as in Figure~\ref{fig:overall}) is trained to reconstruct data from both domains.
The shared latent space (also 2-dimensional) has domain overlap because of the SWD penalty, and class separation due to the linear classifier (decision boundary shown).
This enables bidirectional domain transfer that preserves local structure (shown by the color gradient of datapoints) and class separation (learning to rotate rather than shift and squash).
}
\label{fig:synthetic-one}
\end{figure*}

Modularity enables general and efficient problem solving by recombining predefined components in new ways. As such, modular design is essential to such core pursuits as proving theorems, designing algorithms, and software engineering. Despite the many advances of deep learning, including impressive generative models such as Variational Autoencoders (VAEs) and Generative Adversarial Networks (GANs), end-to-end optimization inhibits modular design by providing no straight-forward way to combine predefined modules. While performance benefits still exist for scaling individual models, such as the recent BigGAN model~\citep{brock2018large}, it becomes increasingly infeasible to retrain such large architectures for every new use case. By analogy, this would be equivalent to having to rewrite an assembly compiler every time one wanted to write a new web application.

Transfer learning and fine-tuning are common methods to reuse individual pretrained modules for new tasks~\citep{bert2018, Liu_2015_ICCV}, but there is still no clear method to get the combinatorial benefits of integrating multiple modules. Here, we explore one such approach by bridging pretrained latent generative models to perform cross-modal domain transfer. 


For cross-modal domain transfer, we seek to train a model capable of transferring instances from a source domain ($x_1$) to a target domain ($x_2$), such that local variations in source domain are transferred to local variations in the target domain. We refer to this property as \emph{locality}. Thus, local interpolation in the source domain would ideally be similar to local interpolation in target domain when transferred.

There are often many possible alignments of semantic attributes that could maintain locality. For instance, absent additional context, there is no reason that dataset images and spoken utterances of the digit ``0'' should align with each other. There may also be no agreed common semantics, like for example between images of landscapes and passages of music, and it is at the liberty of the user to define such connections based on their own intent. Our goal in modeling is to respect such intent and make sure that the correct attributes are connected between the two domains. 

We refer to this property as \emph{semantic alignment}. A user can thus sort a set of data points from in each domain into common bins, which we can use to constrain the cross-domain alignment. We can quantitatively measure the degree of semantic alignment by using a classifier to label transformed data and measuring the percentage of data points that fall into the same bin for the source and target domain. Our goal can thus be stated as learning transformations that preserve locality and semantic alignment, while requiring as few labels from a user as possible. 


To achieve this goal and tackle prior limitations, we propose to abstract the domains with independent latent variable models, and then learn to transfer between the latent spaces of those models. Our main contributions include:

\begin{itemize}
    \item We propose a shared "bridging" VAE to transfer between latent generative models. Locality and semantic alignment of transformations are encouraged by applying a sliced-wasserstein distance (SWD), and a classification loss respectively to the shared latent space.  
    \item We demonstrate with qualitative and quantitative results that our proposed method enables transfer within a modality (image-to-image), between modalities (image-to-audio), and between generative model types (VAE to GAN). 
    \item We show that decoupling the cost of training from that of the base generative models increases training speed by a factor of $\sim200\times$, even for the relatively small base models examined in this work.
\end{itemize}

\section{Related Work} \label{sec:related-works}

\paragraph{Latent Generative Models:}
Deep latent generative models use an expressive neural network function to convert a tractable latent distribution $p(z)$ into the approximation of a population distribution $p^*(x)$.
Two popular variants include VAEs~\citep{kingma2013auto} and GANs~\citep{goodfellow2014generative}.
GANs are trained with an adversarial classifier while VAEs are trained to maximize a variational approximation through the use of evidence lower bound (ELBO).
These classes of models have been thoroughly investigated in many applications and variants~\citep{NIPS2017_7159,li2017mmd,binkowski2018demystifying} including  conditional generation~\citep{mirza2014conditional},
generation of one domain conditioned on another~\citep{dai2017towards,reed2016generative},
generation of high-quality images~\citep{karras2018progressive}, audio~\citep{engel2018gansynth, engel2017nsynth}, and  music~\citep{roberts2018hierarchical}.


\paragraph{Domain Transfer:}

Deep generative models enable domain transfer by learning a smooth mapping between data domains such that the variations in one domain are reflected in the other.
This has been demonstrated to great effect within a single modality, for example transferring between two different styles of image~\citep{pix2pix2016,CycleGAN2017,li2018unsupervised,li2018twin}, video~\citep{wang2018vid2vid}, or music~\citep{mor2018universal}.
These works have been the basis of interesting creative tools~\citep{pix2pixdemo}, as small changes in the source domain are reflected by comparable intuitive changes in the target domain.  

Despite these successes, this line of work has several limitations.
Supervised techniques such as Pix2Pix~\citep{pix2pix2016} and Vid2Vid~\citep{wang2018vid2vid}, are able to transfer between more distant datasets, but require very dense supervision from large volumes of tightly paired data. 
While latent translation can benefit from additional supervision, it does not require all data to be strongly paired data.
Unsupervised methods such as CycleGAN or its variants~\citep{CycleGAN2017, taigman2016} require the two data domains to be closely related (e.g. horse-to-zebra, face-to-emoji, MNIST-to-SVHN)~\citep{li2018unsupervised}.
This allows the model to focus on transferring local properties like texture and coloring instead of high-level semantics.
\citet{chu2017cyclegan} show that CycleGAN transformations share many similarities with adversarial examples, hiding information about the source domain in near-imperceptible high-frequency variations of the target domain.
Latent translation avoids these issues by abstracting the data domains with pretrained models, allowing them to be significantly different.

Perhaps closest to this work is the UNIT framework~\citep{unit2017, cogan2016}, where a shared latent space is learned jointly with both VAE and GAN losses. 
In a similar spirit, they tie the weights of highest layers of the encoders and decoders to encourage learning a common latent space.
While UNIT is sufficient for image-to-image translation (dog-to-dog, digit-to-digit), this work extends to more diverse data domains by allowing independence of the base generative models and only learning a shared latent space to tie the two together.
Also, while joint training has performance benefits for a single domain transfer task, the modularity of latent translation allows specifying new model combinations without the potentially prohibitive cost of retraining the base models for each new combination.

\paragraph{Transfer Learning:}
Transfer learning and fine-tuning aim to reuse a model trained on a specific task for new tasks. For example, deep classification models trained on the ImageNet dataset~\citep{Russakovsky2015} can transfer their learned features to tasks such as object detection~\citep{NIPS2015_5638} and semantic segmentation~\citep{long2015fully}. Natural language processing has also recently seen significant progress through transfer learning of very large pretrained models~\citep{bert2018}. However, easily combining multiple models together in a modular way is still an unsolved problem. This work explores a step in that direction for deep latent generative models.

\paragraph{Neural Machine Translation:}
Unsupervised neural machine translation (NMT) techniques work by aligning embeddings in two different languages. Many approaches use discriminators to make translated embeddings indistinguishable~\citep{zhang2017adversarial, conneau2017word}, similar to applying CycleGAN in latent space. This work takes that approach as a baseline and expands upon it by learning a shared embedding space for each domain. Backtranslation and anchor words~\citep{backtranslation2018, artetxe2018unsupervised} are promising developments in NMT, and exploring their relevance to latent translation of generative models is an interesting avenue for future work.

\section{Method} \label{sec:method}

\begin{figure*}[t!]
\centering
\includegraphics[width=\textwidth,trim={0 180pt 0 0},clip]{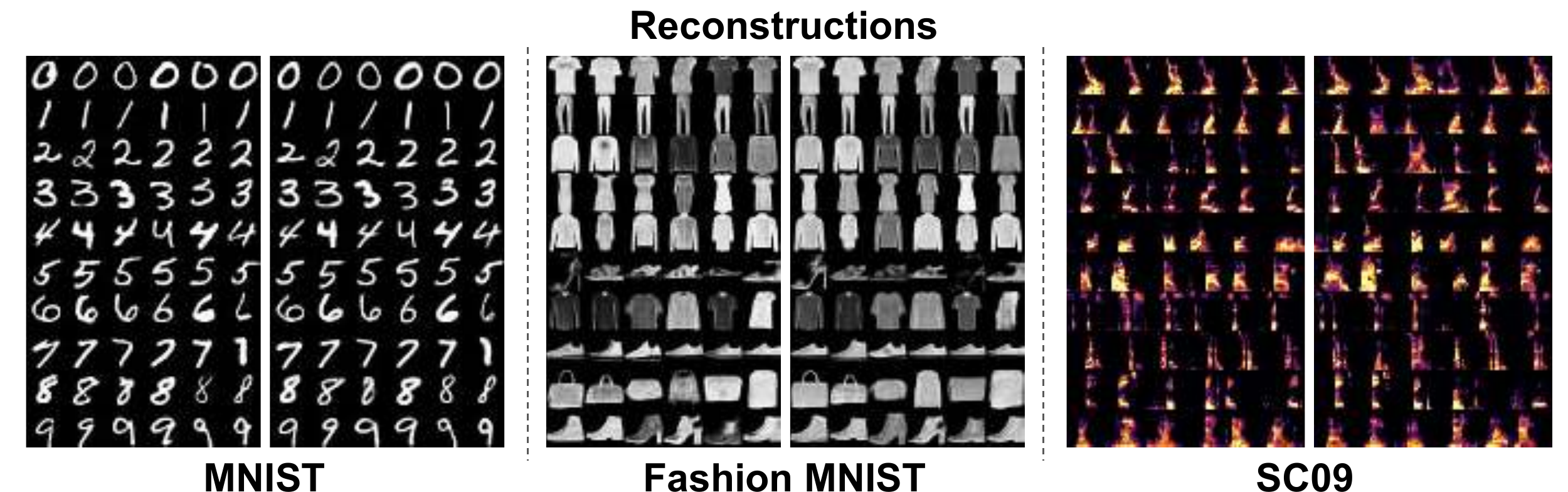}
\caption{Bridging autoencoder reconstructions. 
For each dataset, original data is on the left and reconstructions are on the right. 
For SC09 we show the log magnitude spectrogram of the audio, and label order is the same as MNIST.
The reconstruction quality is limited by the fidelity of the base generative model.
The bridging autoencoder is able to achieve sharp reconstructions because the base models are either VAEs with $\beta<1$ (MNIST, Fashion MNIST) or a GAN (SC09).
More discussions are available in Section~\ref{sec:reconstruction}.
}
\label{fig:reconstruction}
\end{figure*}

Figure~\ref{fig:overall} diagrams our hierarchical approach to latent translation. We start with a separate pretrained generative model, either VAE or GAN, for both the source and target domain. These models give latent embeddings ($z_1$, $z_2$) of the data from both domains ($x_1$, $x_2$). 
For VAEs, data embeddings are given by the encoder, $z \sim q(z|x)$, where $q$ is an encoder network.
Since GANs lack an encoder, we choose latent samples from the prior, $z \sim p(z)$, and then use rejection sampling to keep only samples whose associated data, $x = g(z)$, where $g$ is a generator network, is classified with high confidence by an auxiliary attribute classifier.
We then train a single "bridging" VAE to create a shared latent space ($z'$) that corresponds to both domains.
The bridging VAE shares weights between both latent domains to encourage the model to seek common structure, but we also find it helpful to condition both the encoder $q_{shared}(z'| z, D)$ and decoder $g_{shared}(z| z', D)$, with an additional one-hot domain label, $D$, to allow the model some flexibility to adapt to variations particular to each domain.
While the base VAEs and GANs have spherical Gaussian priors, we penalize the KL-Divergence term for VAEs to be less than 1 (also known as a $\beta$-VAE~\citep{higgins2016beta}), allowing the models to achieve better reconstructions and retain some structure of the original dataset for the bridging VAE to model. Full architecture and training details are available in the Supplementary Material. 

The domain conditional bridging VAE objective consists of three loss terms:
\begin{enumerate}
    \item \textbf{Evidence Lower Bound (ELBO)}. Standard VAE loss. For each domain $d \in \{1, 2\}$, 
    \begin{equation*}
    \begin{split}
        \gL_{d}^{\mathrm{ELBO}} = & - \E_{z' \sim Z_d' } \left[ \log \pi (z_d; g(z', D=d) \right] \\
                                  & + \beta_{\mathrm{KL}} \KL \left( q(z'|z_d, D=d) \Vert p(z') \right)
    \end{split}
    \end{equation*}
    where the likelihood $\pi (z; g) $  is a spherical Gaussian $\gN(z; g, \sigma^2 I)$, and $\sigma$ and $\beta_{\mathrm{KL}}$ are hyperparmeters set to 1 and 0.1 respectively to encourage reconstruction accuracy. 
    
    \item  \textbf{Sliced Wasserstein Distance (SWD)}~\citep{bonneel2015sliced}. The distribution distance between mini-batches of samples from each domain in the shared latent space $(z_1', z_2')$,  
    $$\gL^{\mathrm{SWD}} = {1}/{|\Omega |} \sum_{\omega \in \Omega} W_2^2 \left( \mathrm{proj}(z_1', \omega), \mathrm{proj}(z_2', \omega) \right) $$
    where $\Omega$ is a set of random unit vectors, $\mathrm{proj}(A, a)$ is the projection of $A$ on vector $a$, and $W_2^2(A,B)$ is the quadratic Wasserstein distance.
    
    \item \textbf{Classification Loss (Cls)}. For each domain $d\in\{1,2\}$, we enforce semantic alignment with attribute labels $y$ and a classification loss in the shared latent space:
    $$ \gL_{d}^{\mathrm{Cls}} = \E_{z'\in Z_d'} H( f(z'), y ) $$ 
    where $H$ is the cross entropy loss, $f(z')$ is a linear classifier. 
\end{enumerate}
Including terms for both domains, it gives the total training loss,
$$ \gL = (\gL_{1}^{\mathrm{ELBO}} + \gL_{2}^{\mathrm{ELBO}})  + \beta_{\mathrm{SWD}} \gL^{\mathrm{SWD}} + \beta_{\mathrm{Cls}} (\gL_{1}^{\mathrm{Cls}} + \gL_{2}^{\mathrm{Cls}}) $$
where $\beta_{\mathrm{SWD}}$ and $\beta_{\mathrm{Cls}}$ are scalar loss weights. The transfer procedure is illustrated Figure~\ref{fig:synthetic-one} using synthetic data. For reconstructions, data $x_1$ is passed through two encoders, $z_1 \sim q(z_1 | x_1)$, $z' \sim q_{\mathrm{shared}}(z' | z_1, D=1)$, and two decoders, $\hat{z_1} \sim g_{\mathrm{shared}}(\hat{z_1} | z', D=1)$, $\hat{x_1} \sim g(\hat{x_1} | \hat{z_1})$. For transformations, the encoding is the same, but decoding uses decoders (and conditioning) from the second domain, $\hat{z_2} \sim g_{\mathrm{shared}}(\hat{z_2} | z', D=2)$, $\hat{x_2} \sim g(\hat{x_2} | \hat{z_2})$. Further analysis and intuition behind loss terms is given in Figure~\ref{fig:synthetic-one}.

\begin{figure*}[t!]
\centering
\includegraphics[width=0.96\textwidth,trim={0 180pt 0 0},clip]{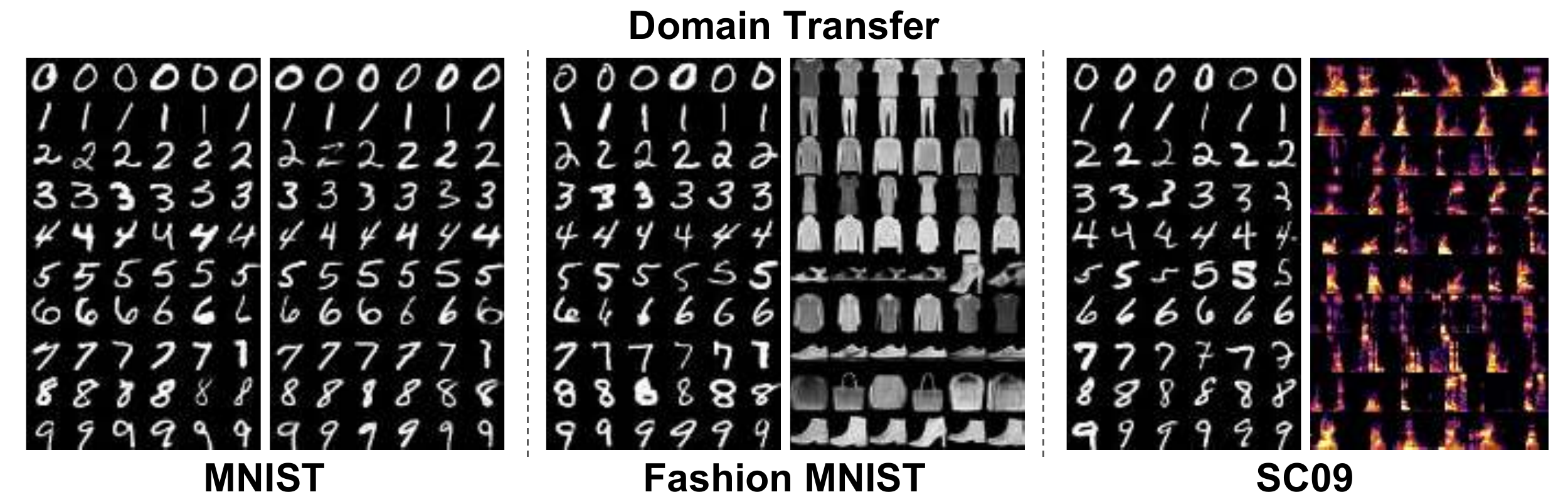}
\caption{
Domain Transfer from an MNIST VAE to a separately trained MNIST VAE, Fashion MNIST VAE, and SC09 GAN.
For each dataset, the left is the data from the source domain and on the right is transformations to the target domain.
Domain transfer maintains the label identity, and a diversity of outputs, mapping local variations in the source domain to local variations in the target domain.
}
\label{fig:transfer}
\end{figure*}

\begin{table*}[t]
\centering
\small
\resizebox{.90\linewidth}{!}{
    \begin{tabular}{c c c c c c | c } 
     \toprule
     Transfer Accuracy &  &  & & & & FID \\
     \midrule
      & MNIST $\to $ & MNIST $\to$   & F-MNIST & MNIST  & SC09 $\to$  &  MNIST $\to$ \\  
     Model Type & MNIST      & F-MNIST   & $\to$ MNIST   & $\to$ SC09 & MNIST & F-MNIST \\ 
     \midrule
     Pix2Pix & - & $0.77$ & $0.08$ & $\times$ & $\times$ & $0.087$ \eat{$0.079$} \\
     CycleGAN & - & $0.08$ & $0.13$ & $\times$ & $\times$ & $0.361$ \eat{$0.333$} \\ 
     Latent CycleGAN & $0.08$ & $0.10$  &  $0.10$ & $0.09$ & $0.11$ & $0.081$ \\
     This work & $\mathbf{0.98}$ & $\mathbf{0.95}$ & $\mathbf{0.89}$ & $\mathbf{0.67}$ & $\mathbf{0.98}$ & $\mathbf{0.004}$ \eat{$\mathbf{0.006}$} \\ 
     \bottomrule
    \end{tabular}
}
\caption{
Quantitative comparison of domain transfer accuracy and quality.
We compare to preexisting approaches,  Pix2Pix \citep{pix2pix2016} and CycleGAN \citep{CycleGAN2017} trained on raw-pixels, and latent tranlsation via CycleGAN (Latent CycleGAN).
All baseline models are trained with pairs of class-aligned data.
As in Figure~\ref{fig:transfer}, MNIST $\to$ MNIST transfers between pretrained models with different initial conditions.
Pix2Pix and CycleGAN fail to train on MNIST $\to$ SC09, as the two domains are too distinct.
We compute class accuracies and Fr\'{e}chet Inception Distance (FID, lower value indicates a better image quality), using pretrained classifiers on the target domain.
Latent CycleGAN accuracies are similar to chance because the cyclic reconstruction cost dominates and encourages learning the identity function.
}
\label{table:transfer-accuarcy}
\end{table*}

\begin{figure*}[t!]
\centering
\includegraphics[width=0.96\textwidth,trim={0 140pt 0 0},clip]{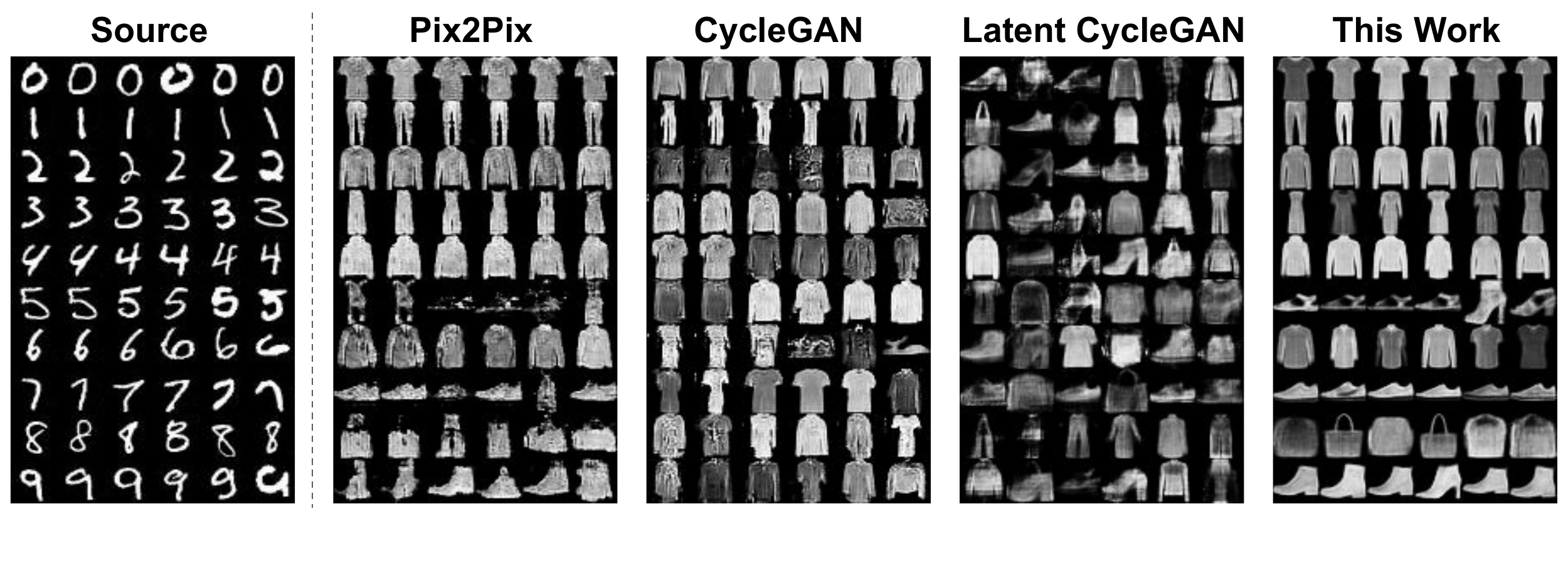}
\caption{Qualitative comparison of domain transfer across models, as in the last column of Table~\ref{table:transfer-accuarcy}. Pix2Pix seems to have collapsed to a prototype for each class and CycleGAN has collapsed to output mostly shirts and jackets. Latent CycleGAN also has no class structure, as compared to the bridging autoencoder that generates clear and diverse class-appropriate transformations.\\
\centerline{(More discussions for Figure~\ref{fig:transfer}, Table~\ref{table:transfer-accuarcy} and Figure~\ref{fig:failed-case-for-existing-apparoches} are available in Section~\ref{sec:domain-transfer})}
}
\label{fig:failed-case-for-existing-apparoches}
\end{figure*}

\section{Experiments} \label{sec:experiment}

\subsection{Datasets}

While the end goal of our method is to enable mapping between arbitrary datasets, for quantitative evaluation we restrict ourselves to three domains where there exist a somewhat natural alignment for comparison:
\begin{enumerate}
\item \textbf{MNIST}~\citep{lecun1998mnist}, which contains images of hand-written digits of 10 classes from ``0'' to ``9''.
\item \textbf{Fashion MNIST}~\citep{xiao2017/online}, which contains fashion related objects such as shoes, t-shirts, categorized into 10 classes. The structure of data and the size of images are identical to MNIST.
\item \textbf{SC09}, a subset of the Speech Commands Dataset~\footnote{Dataset available at \url{https://ai.googleblog.com/2017/08/launching-speech-commands-dataset.html}}, which contains spoken digits from ``0'' to ``'9''.
This is a much noisier and more difficult dataset than the others, with 16,000 dimensions (1 second 16kHz) instead of 768. 
Since WaveGAN lacks an encoder, the bridging VAE dataset is composed of samples from the prior rather than encodings of the data.
We collect 1,300 prior samples per a label by rejecting samples with a maximum softmax output $< 0.95$ on the pretrained WaveGAN classifier for spoken digits. 
\end{enumerate}

For MNIST and Fashion MNIST, we pretrain VAEs with MLP encoders and decoders following \citet{engel2018latent}. 
For SC09, we chose to use the publicly available WaveGAN~\citep{donahue2018adversarial}\footnote{Code and pretrained model available at \url{https://github.com/chrisdonahue/wavegan}} because we wanted a global latent variable for the full waveform (as opposed to a distributed latent code as in \citet{engel2017nsynth}) and VAEs fail on this more difficult dataset. 
It also gives us the opportunity to explore transferring between different classes of models. For the bridging VAE, we also use stacks of fully-connected linear layers with ReLU activation and a final ``gated mixing layer''.
Full network architecture details are available in the Supplementary Material.
We would like to emphasize that we only use \emph{class level} supervision for enforcing semantic alignment with the latent classifier.

\begin{figure*}[t!]
\centering
\includegraphics[width=\textwidth,trim={0 115pt 0 1pt},clip]{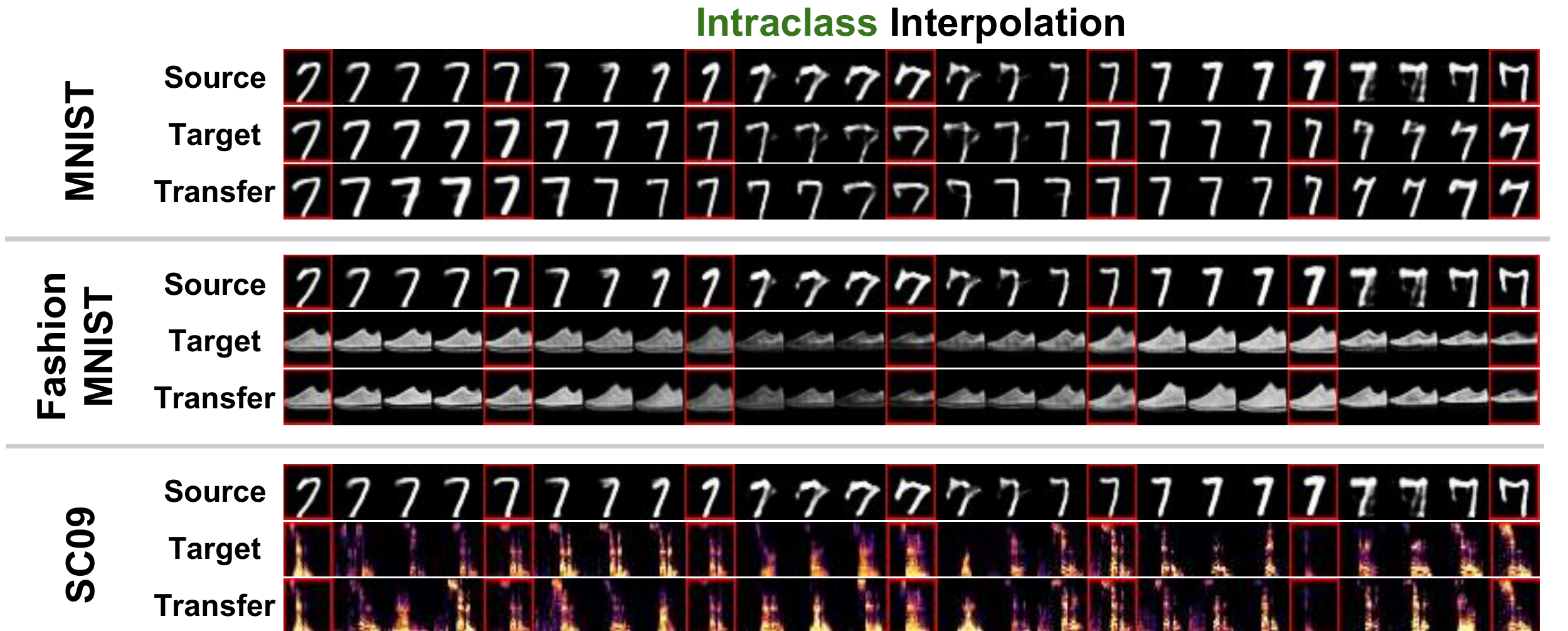}
\caption{
Interpolation within a class using bridging autoencoders. 
Columns of images in red squares are fixed points, with three rows of interpolations.
(1) Source: Interpolate in source domain between fixed data points,
(2) Target: Transfer fixed data points in source domain to target domain and interpolate between transferred fixed points there, 
(3) Transfer: Transfer all points in first row to the target domain.
Note that Transfer interpolation produces smooth variation of data attributes like Target interpolation, indicating that local variations in the source domain are mapped to local variations in the target domain.
More discussions are available in Section~\ref{sec:interpolation}.
}
\label{fig:intra-class-interpolation}
\end{figure*}

We examine three scenarios of domain transfer:
\begin{enumerate}
\item MNIST $\leftrightarrow$ MNIST. We first train two lower-level VAEs from different initial conditions. The bridging autoencoder is then tasked with transferring between latent spaces while maintaining the digit class from source to target. 
\item MNIST $\leftrightarrow$ Fashion MNIST. In this scenario, we specify a global one-to-one mapping between 10 digit classes and 10 fashion object classes (available in the Supplementary Material)
The bridging autoencoder is tasked with preserving this mapping as it transfers between images of digits and clothing. 
\item MNIST $\leftrightarrow$ SC09. For the speech dataset, we first train a GAN to generate audio waveforms \citep{donahue2018adversarial} of spoken digits.   The bridging autoencoder is then tasked with transferring between a VAE of written digits and a GAN of spoken digits.\footnote{
Generated audio samples available in the supplemental material and at
\url{https://drive.google.com/drive/folders/1mcke0IhLucWtlzRchdAleJxcaHS7R-Iu}
}
\end{enumerate}

\subsection{Baselines}
\label{sec:baselines}

Where possible, we compare latent translation with a bridging VAE to three existing approaches.
As a straightforward baseline, we train Pix2Pix~\citep{pix2pix2016} and CycleGAN~\citep{CycleGAN2017} models to perform domain transfer directly in the data space. 
In analogy to the NMT literature, we also provide a baseline of latent translation with a CycleGAN in latent space (Latent CycleGAN).
To encourage semantic alignment in baselines, we provide additional supervision by only training on class aligned pairs of data.

\section{Results} \label{sec:results}

\subsection{Reconstruction}
\label{sec:reconstruction}

For bridging VAEs, qualitative reconstruction results are shown in Figure~\ref{fig:reconstruction}.
The quality is limited by the fidelity of the base generative model, which is quite sharp for MNIST and Fashion MNIST (VAEs with $\beta<1$).
However, despite being accurate and intelligible, the "ground truth" of WaveGAN samples for SC09 is quite noisy, as the model is not completely successful at capturing this relatively difficult dataset. 
For all datasets \eat{, it is clear that } the bridging VAE produces reconstructions of comparable quality to the original data.

Quantitatively, we can calculate reconstruction accuracies as shown in Table~\ref{table:reconstruction-accuarcy}. 
With pretrained classifiers in each data domain, we measure the percentage of reconstructions that do not change in their predicted class.
As expected from the qualitative results, the bridging VAE reconstruction accuracies for MNIST and Fashion MNIST are similar to the base models, indicating that it has learned the latent data manifold.

The lower accuracy on SC09 likely reflects the worse performance of the base generative model on the difficult dataset, leading to greater variance in the latent space.
For example, only $31.8\%$ samples from the GAN prior give the classifier high enough confidence to be used for training the bridging VAE. Despite this, the model still achieves relatively high reconstruction accuracy.

\begin{table}[h]
\centering
\resizebox{.9\columnwidth}{!}{
    \begin{tabular}{c c c c} 
     \toprule
      Accuracy & MNIST & Fashion MNIST & SC09 \\
     \midrule
     Base Model & $0.995$ & $0.952$ & - \\  
     Bridging VAE & $0.989$ & $0.903$ & $0.739$ \\ 
     \bottomrule
    \end{tabular}
}
\caption{Reconstruction accuracy for base generative models and bridging VAE. WaveGAN does not have an encoder, and thus does not have a base reconstruction accuracy.
}
\label{table:reconstruction-accuarcy}
\end{table}

\subsection{Domain Transfer}
\label{sec:domain-transfer}
For bridging VAEs, qualitative transfer results are shown in Figure~\ref{fig:transfer}. 
For each pair of datasets, domain transfer maintains the label identity, and a diversity of outputs, at a similar quality to the reconstructions in Figure~\ref{fig:reconstruction}. 

Quantitative results are given in Table~\ref{table:transfer-accuarcy}. Given that the datasets have pre-aligned classes, when evaluating transferring from data $x_{d_1}$ to $x_{d_2}$, the reported accuracy is the portion of instances that $x_{d_1}$ and $x_{d_2}$ have the same predicted class.
We also compute Fr\'{e}chet Inception Distance (FID) as a measure of image quality using features form the same pretrained classifiers~\citep{heusel2017gans}.
The bridging VAE achieves very high transfer accuracies (bottom row), which are comparable in each case to reconstruction accuracies in the target domain. 
Interestingly, it follows that MNIST $\to$ SC09 has lower accuracy than SC09 $\to$ MNIST, reflecting the reduced quality of the WaveGAN latent space.

In Table~\ref{table:transfer-accuarcy} we also quantitatively compare with the baseline models where possible. We exclude Pix2Pix and CycleGAN from MNIST $\to$ MNIST as it involves transfer between different initializations of the same model on the same dataset and does not apply. Despite a large hyperparameter search,  Pix2Pix and CycleGAN fail to train on MNIST $\leftrightarrow$ SC09, as the two domains are too distinct. Qualitative results for MNIST $\to$ Fashion MNIST are shown in Figure~\ref{fig:failed-case-for-existing-apparoches}. Pix2Pix has higher accuracy than the other baselines because it seems to have collapsed to a prototype for each class, while CycleGAN has collapsed to outputs mostly shirts and jackets. In contrast, the bridging autoencoder generates diverse and class-appropriate transformations, with higher image quality which is reflected in their lower FID scores.

Finally, we found that latent CycleGAN had transfer accuracies roughly equal to chance. 
Unlike NMT, the base latent spaces have spherical Gaussian priors that make the cyclic reconstruction loss easy to optimize, encouraging generators to learn an identity function. 
Indeed, the samples in Figure~\ref{fig:failed-case-for-existing-apparoches} resemble samples from the Gaussian prior, which are randomly distributed in class. 
This motivates the need for new techniques for latent translation such as the bridging VAE in this work.

\begin{figure*}[t!]
\centering
\includegraphics[width=\textwidth,trim={0 115pt 0 1pt},clip]{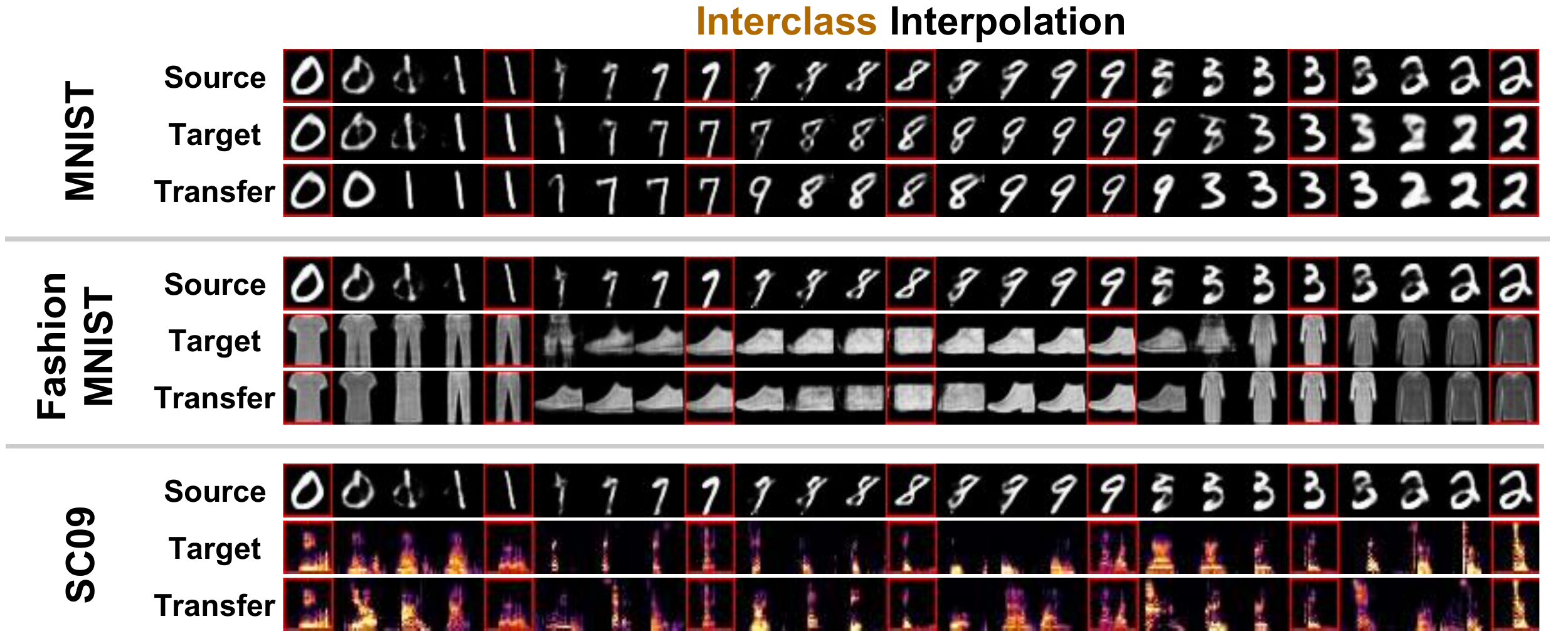}
\caption{
Interpolation between classes. The arrangement of images is the same as Figure~\ref{fig:intra-class-interpolation}.
Source and Target interpolation produce varied, but sometimes unrealistic, outputs between class fixed points.
The bridging autoencoder is trained to stay close to the marginal posterior of the data distribution. As a result Transfer interpolation varies smoothly within a class but makes larger jumps at class boundaries to avoid unrealistic outputs.
More discussions are available in Section~\ref{sec:interpolation}.
}
\label{fig:inter-clas-interpolation}
\end{figure*}

\subsection{Interpolation}
\label{sec:interpolation}
Interpolation can act as a good proxy for locality, as good interpolation requires small changes in the source domain to cause small changes in the target domain.
We show intraclass and interclass interpolation in Figure~\ref{fig:intra-class-interpolation} and Figure~\ref{fig:inter-clas-interpolation} respectively.
We use spherical interpolation~\citep{slerp},
as we are interpolating in a Gaussian latent space.
The figures compare three types of interpolation:
(1) the interpolation in the source domain's latent space, which acts a baseline for smoothness of interpolation for a pretrained generative model,
(2) transfer fixed points to the target domain's latent space and interpolate in that space, and
(3) transfer all points of the source interpolation to the target domain's latent space, which shows how the transferring warps the latent space.

Note for intraclass interpolation, the second and third rows have comparably smooth trajectories, reflecting that locality has been preserved. 
For interclass interpolation in Figure~\ref{fig:inter-clas-interpolation} interpolation is smooth within a class, but between classes the second row blurs pixels to create blurry combinations of digits, while the full transformation in the third row makes sudden transitions between classes. This is expected from our training procedure as the bridging autoencoder is modeling the marginal posterior of each latent space, and thus always stays on the manifold of the actual data during interpolation.

\subsection{Efficiency and Ablation Analysis}
\label{efficiency-and-ablation-analysis}

Since our method is a semi-supervised method,
we want to know how effectively our method leverages the labeled data.
In Table~\ref{table:data-efficiency} we show for the MNIST $\to$ MNIST setting
the performance measured by transfer accuracy with respect to the number of labeled data points. Labels are distributed evenly among classes. The accuracy of transformations grows monotonically with the number of labels and reaches over 50\% with as few as 10 labels per a class. Without labels, we also observe accuracies greater than chance due to unsupervised alignment introduced by the SWD penalty in the shared latent space. 

\begin{table}[h!]
\centering
\resizebox{.95\columnwidth}{!}{
    \begin{tabular}{c c c c c c c} 
     \toprule
     \ Labels / Class  & $0$ & $1$ & $10$ & $100$ & $1000$ & $6000$ \\
     \midrule
     Accuracy & $0.1390$ & $0.339$ & $0.524$ &$0.6810$ & $0.898$ & $0.980$ \\ 
     \bottomrule
    \end{tabular}
}
\caption{MNIST $\to$ MNIST transfer accuracy with labeled data.}
\label{table:data-efficiency}
\end{table}

Besides data efficiency, pretraining the base generative models has computational advantages.
For large generative models that take weeks to train, it would be infeasible to retrain the entire model for each new cross-domain mapping. The bridging autoencoder avoids this cost by only retraining the latent transfer mappings. As an example from these experiments, training the bridging autoencoder for MNIST $\leftrightarrow$ SC09 takes about half an hour on a single GPU, while retraining the SC09 WaveGAN takes around four days.

Finally, we perform an ablation study to confirm the benefits of each architecture component to transfer accuracy. For consistency, we stick to the MNIST $\to$ MNIST setting with fully labeled data. In Table~\ref{table:conponent-contribution}, we see that the largest contribution to performance is from the domain conditioning signal, allowing the model to adapt to the specific structure of each domain. Further, the increased overlap in the shared latent space induced by the SWD penalty is reflected in the greater transfer accuracies. 

\begin{table}[h!]
\centering
\resizebox{.80\columnwidth}{!}{
    \begin{tabular}{c c c c} 
     \toprule
     \makecell{Method \\ Ablation}  & \makecell{Unconditional \\ VAE} & \makecell{Domain \\ Conditional \\ VAE} & \makecell{Domain \\ Conditional \\ VAE + SWD} \\
     \midrule
     Accuracy & $0.149$ & $0.849$ & $0.980$ \\ 
     \bottomrule
    \end{tabular}
}
\caption{Ablation study of MNIST $\to$ MNIST transfer accuracies.}
\label{table:conponent-contribution}
\end{table}

\section{Conclusion} \label{sec:conclusion}

We have demonstrated an approach to learn mappings between disparate domains by bridging the latent codes of each domain with a shared autoencoder. We find bridging VAEs are able to achieve high transfer accuracies, smoothly map interpolations between domains, and even connect different model types (VAEs and GANs). Here, we have restricted ourselves to datasets with intuitive class-level mappings for the purpose of quantitative comparisons, however there are many interesting creative possibilities to apply these techniques between domains without a clear semantic alignment. Finally, as a step towards modular design, we combine two pretrained models to solve a new task for significantly less computational resources than retraining the models from scratch.

\bibliography{ref}
\bibliographystyle{icml2019}


\clearpage
\appendix
\appendixpage

\begin{figure*}[h!]
	\centering
	\includegraphics[width=0.9\textwidth,trim={0 50pt 0 0},clip]{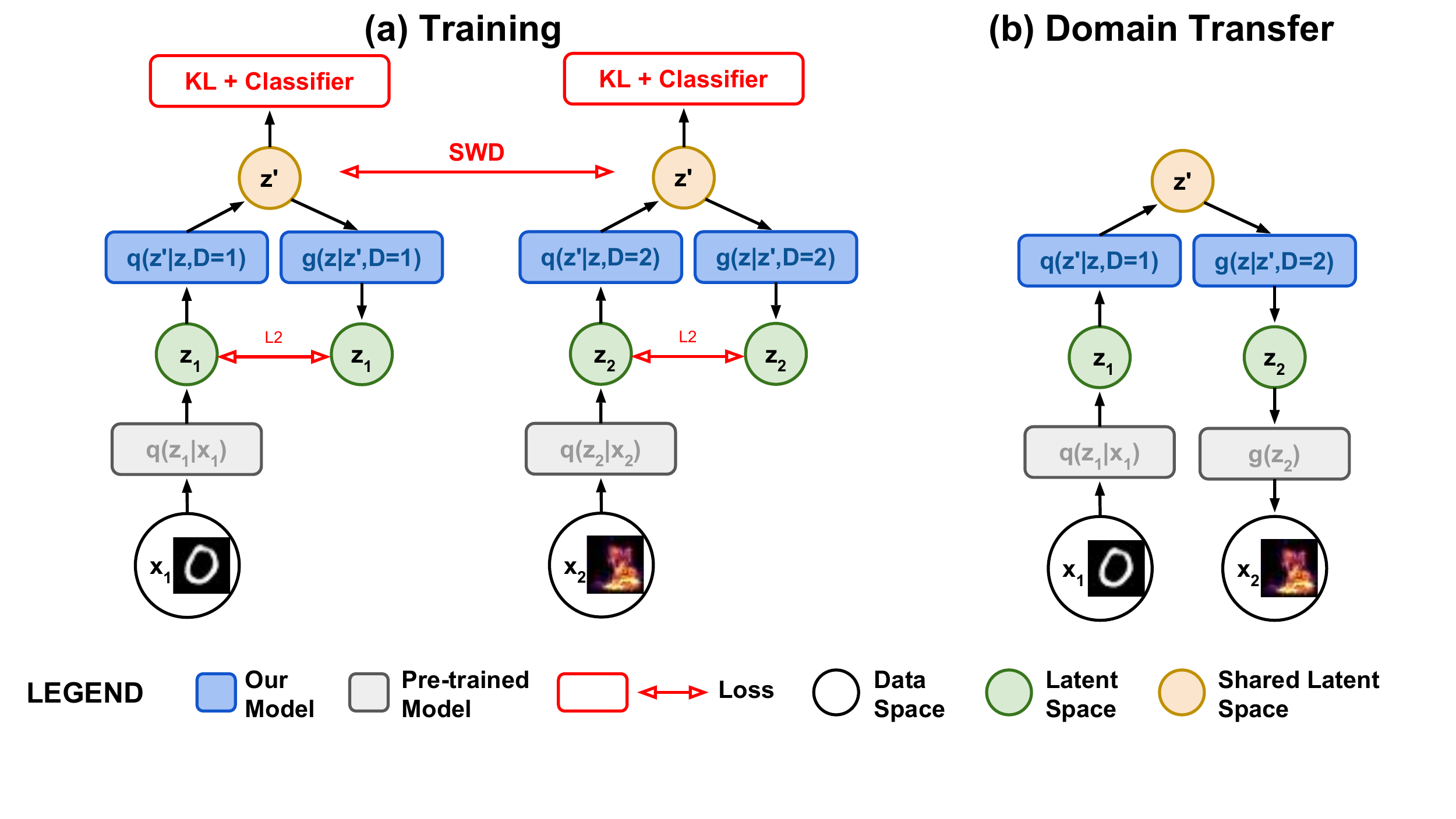}
	\caption{Architecture and training. Architecture and training. Our method aims at transfer from one domain to another domain such that the correct semantics (e.g., label) is maintained across domains and local changes in the source domain should be reflected in the target domain.
		To achieve this, we train a model to transfer between the latent spaces of pre-trained generative models on source and target domains.
		\textbf{(a)} The training is done with three types of loss functions:
		(1) The VAE ELBO losses to encourage modeling of $z_1$ and $z_2$, which are denoted as L2 and KL in the figure.
		(2) The Sliced Wasserstein Distance loss to encourage cross-domain overlapping in the shared latent space, which is denoted as SWD.
		(3) The classification loss to encourage intraclass overlap in the shared latent space, which is denoted as Classifier.
		The training is semi-supervised, since (1) and (2) requires no supervision (classes) while only (3) needs such information.
		\textbf{(b)} To transfer data from one domain $x_1$ (an image of digit ``0'') to another domain $x_2$ (an audio of human saying ``zero'', shown in form of spectrum in the example),
		we first encode $x_1$ to $z_1 \sim q(z_1 | x_1)$,
		which we then further encode to a shared latent vector $z'$ using our conditional encoder, $z' \sim q(z' | z_1 , D=1)$, where $D$ donates the operating domain.
		We then decode to the latent space of the target domain $z_2 = g(z|z', D=2)$ using our conditional decoder, which finally is used to generate the transferred audio $x_2 = g(x_2| z_2)$.}
	\label{fig:overall-full}
\end{figure*}

\section{Training Target Design} \label{sec:training-design}

We want to archive following three goals for the proposed VAE for latent spaces:

\begin{enumerate}
	\item It should be able to model the latent space of both domains, including modeling local changes as well. 
	\item It should encode two latent spaces in a way to enable domain transferability. This means encoded $z_1$ and $z_2$ in the shared latent space should occupy overlapped spaces.
	\item The transfer should be kept in the same class. That means, regardless of domain, $z$s for the same class should occupy overlapped spaces.
\end{enumerate}
With these goals in mind, we propose to use an optimization target composing of three kinds of losses.
In the following text for notational convenience, 
we denote approximated posterior $Z_d' \triangleq q(z' | z_d, D=d), z_d \sim q(z_d|x_d), x_d\sim p(x_d)$ for $d\in\{1,2\}$, the process of sampling $z_d'$ from domain $d$.

\begin{figure*}[t!]
	\centering
	\includegraphics[width=\textwidth,trim={20pt 0 40pt 0},clip]{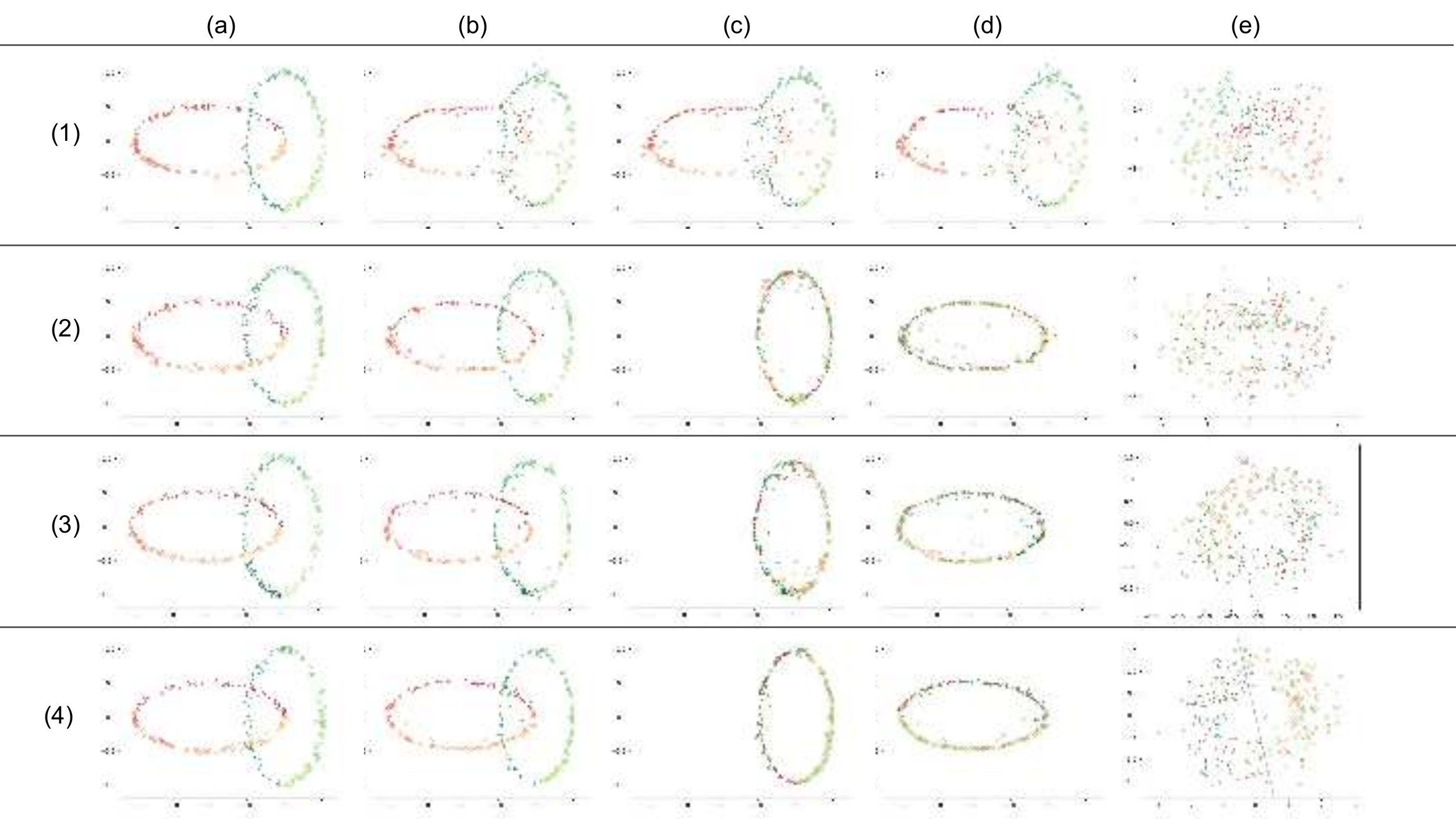}
	\caption{Synthetic data to demonstrate the transfer between 2-D latent spaces with 2-D shared latent space.
		Better viewed with color and magnifier.
		Columns (a) - (e) are 
		synthetic data in latent space,
		reconstructed latent space points using VAE,
		domain 1 transferred to domain 2,
		domain 2 transferred to domain 1,
		shared latent space, 
		respectively,
		follow the same arrangement as Figure~\ref{fig:synthetic-one}.\
		Each row represent a combination of our proposed components as follows:
		\textbf{(1)} Regular, unconditional VAE. Here transfer fails and the shared latent space are divided into region for two domains.
		\textbf{(2)} Conditional VAE. Here exists an overlapped shared latent space. However the shared latent space are not mixed well.
		\textbf{(3)} Conditional VAE + SWD. Here the shared latent space are well mixed, preserving the local changes across domain transfer.
		\textbf{(4)} Conditional + SWD + Classification. This is the best scenario that enables both domain transfer and class preservation as well as local changes. An overall observation is that each proposed component contributes positively to the performance in this synthetic data, which serves as a motivation for our decision to include all of them.
	}
	\label{fig:synthetic-full}
\end{figure*}

For reference, 
In Figure~\ref{fig:synthetic-full}
we show the intuition to design and the contribution to performance from each loss terms.
Also, the complete diagram of traing target is detailed in Figure~\ref{fig:overall-full}.

\paragraph{1. Modeling two latent spaces with local changes.} 
VAEs are often used to model data with local changes in mind, usually demonstrated with smooth interpolation,
and we believe this property also applies when modeling the latent space of data.
Consider for each domain $d \in \{1, 2\}$, the VAE is fit to data to maximize the ELBO (Evidence Lower Bound)    \begin{equation*}
\begin{split}
\gL_{d}^{\mathrm{ELBO}} = & - \E_{z' \sim Z_d' } \left[ \log \pi (z_d; g(z', D=d) \right] \\
& + \beta_{\mathrm{KL}} \KL \left( q(z'|z_d, D=d) \Vert p(z') \right)
\end{split}
\end{equation*}
where both $q$ and $g$ are fit to maximize $\gL_{d}^{\mathrm{ELBO}}$.
Notably, the latent space $z$s are continuous, so we choose the likelihood $\pi (z; g) $ to be the product of $\gN(z; g, \sigma^2 I)$, where we set $\sigma$ to be a constant that
effectively sets $\log \pi (z; g) = || z - g ||_2 $, which is the L2 loss in Figure~\ref{fig:overall-full} (a).
Also,  $\KL$ is denoted as KL loss in Figure~\ref{fig:overall-full} (a).

\paragraph{2. Cross-domain overlapping in shared latent space.}
Formally, we propose to measure the cross-domain overlapping through the distance between following two distributions as a proxy:
the distribution of $z'$ from source domain (e.g., $z_1' \sim Z_1'$) and that from the target domain  (e.g., $z_2' \sim Z_1'$).
We use Wasserstein Distance~\citep{arjovsky2017wasserstein}  to measure the distance of two sets of samples (this notion straightforwardly applies to the mini-batch setting) $S_1'$ and $S_2'$,
where $S_1'$ is sampled from the source domain $z_1' \sim Z_1'$ and  $S_1'$ from the target domain $z_2' \sim Z_d'$.
For computational efficiency and  inspired by \cite{deshpande2018generative}, we use SWD, or Sliced Wasserstein Distance~\citep{bonneel2015sliced} between $S_1'$ and $S_2'$ as a loss term to encourage cross-domain overlapping in shared latent space.
This means in practice we introduce the loss term
\[ \gL^{\mathrm{SWD}} = \frac{1}{|\Omega |} \sum_{\omega \in \Omega} W_2^2 \left( \mathrm{proj}(S_1', \omega), \mathrm{proj}(S_2', \omega) \right)  \]
where $\Omega$ is a set of random unit vectors, $\mathrm{proj}(A, a)$ is the projection of $A$ on vector $a$,
and $W_2^2(A,B)$ is the quadratic Wasserstein distance, which in the one-dimensional case can be easily solved by monotonically pairing points in $A$ and $B$,
as proven in \cite{deshpande2018generative}.

\paragraph{3. Intra-class overlapping in shared latent space.}
We want that regardless of domain, $z$s for the same class should occupy overlapped spaces, 
so that instance of a particular class should retain its label through the transferring.
We therefore introduce the following loss term for both domain $d\in\{1,2\}$
\[ \gL_{d}^{\mathrm{Cls}} = \E_{z'\in Z_d'} H( f(z'), l_{x'} ) \]
where H is the cross entropy loss, $f(z')$ is a one-layer linear classifier, and $l_{x'}$ is the one-hot representation of label of $x'$ where $x'$ is the data associated with $z'$. 
We intentionally make classifier $f$ as simple as possible in order to encourage more capacity in the VAE instead of the classifier.
Notably, unlike previous two categories of losses that are unsupervised, this loss requires labels and is thus supervised.

\section{Model Architecture} \label{sec:network-architecture-our-models}

\subsection{Bridging VAE}

\begin{figure*}[t!]
	\centering
	\includegraphics[width=\textwidth,trim={0 0 0 0},clip]{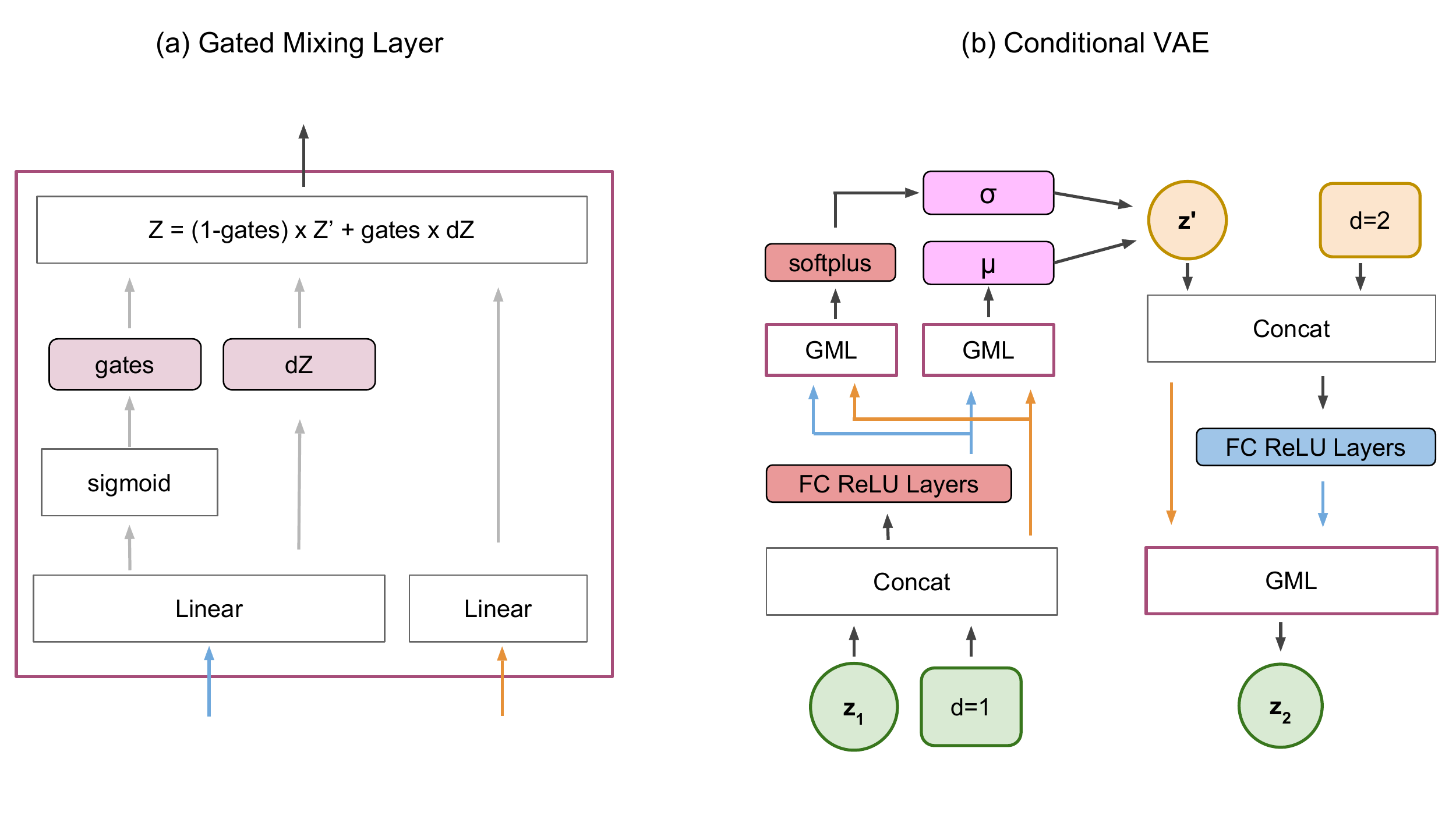}
	\caption{Model Architecture for our Conditional VAE.
		\textbf{(a)} Gated Mixing Layer, or GML, as an important building component.
		\textbf{(b)} Our conditional VAE with GML.
	}
	\label{fig:network-architecture-our-models}
\end{figure*}

The model architecture of our proposed Bridging VAE is illustrated in Figure~\ref{fig:network-architecture-our-models}.
The model relies on Gated Mixing Layers, or GML.
We find empirically that GML improves performance by a large margin than linear layers,
for which we hypothesize that this is because both the latent space ($z_1, z_2$) and the shared latent space $z'$ are Gaussian space,
GML helps optimization by starting with a good initialization.
We also explore other popular network components such as residual network and batch normalization, but find that they are not providing performance improvements.
Also, the condition is fed to encoder and decoder as a 2-length one hot vector indicating one of two domains.

For all settings, we use the dimension of latent space  $100$,
$\beta_{\mathrm{SWD}}=1.0$ and $\beta_{\mathrm{CLs}}=0.05$, 
Specifically, for MNIST $\leftrightarrow$ MNIST and MNIST $\leftrightarrow$ Fashion MNIST, 
we use the dimension of shared latent space $8$, 4 layers of FC (fully Connected Layers) of size $512$ with ReLU activation, $\beta_{\mathrm{KL}}=0.05$, 
$\beta_{\mathrm{SWD}}=1.0$ and $\beta_{\mathrm{CLs}}=0.05$;
while for  MNIST $\leftrightarrow$ SC09,
we use the dimension of shared latent space $16$, 8 layers of FC (fully Connected Layers) of size $1024$ with ReLUa ctivation, $\beta_{\mathrm{KL}}=0.01$, 
$\beta_{\mathrm{SWD}}=3.0$ and $\beta_{\mathrm{CLs}}=0.3$.
The difference is due to that GAN does not provide posterior, so the latent space points estimated by the classifier is much harder to model.

For optimization, we use Adam optimizer\citep{kingma2014adam} with learning rate $0.001$, 
$beta_1=0.9$ and 
$beta_2=0.999$.
We train $50000$ batches with batch size $128$.
We do not employ any other tricks for VAE training.

\subsection{Base Models and Classifiers}

\paragraph{Base Model for MNIST and Fashion-MNIST}
The base model for data space $x$ (in our case MNIST and Fashion-MNIST) is a standard VAE with latent space $z$,
consisting of an encoder function $q(z|x)$ that serves as an approximation to the posterior $p(z|x)$,
a decoder function $g(z)$ and a likelihood  $\pi\left( z; g(z) \right)$ that is defined as $p(z|x)$.
Since we normalize MNIST and Fashion MNIST's pixel values such that they become continuous values in range $[0, 1]$,
The likelihood $\pi\left( z; g \right)$ is accordingly Gaussian $\mathcal{N}\left(z; g,\sigma_{x} I\right)$.
Both $q$ and $g$ are optimized to the Evidence Lower Bound (ELBO):
\[ 
\gL^{\mathrm{ELBO}} = - \E_{x \sim X } \left[ \log \pi (x; g(z) \right] + \beta \KL \left( q(z|x) \Vert p(z) \right) 
\]
where $p(z)$ is a tractable simple prior which is $\mathcal{N}\left(0; I\right)$
We parameterize both encoder and decoder with neural networks.
Specifically, the encoder 
consists of 3 layers of FC (Fully Connected Layers) of size $1024$ with ReLU activation,
on top of which are an affine transformation for $z_\mu$ and an FC with sigmoid activation for $z_\sigma$ (both of size of latent space, which is $100$),
which give $ q(z|x) \triangleq  \mathcal{N}\left(z_\mu; z_\sigma \right) $.
the decoder $g$ consists of 3 layers of FC (Fully Connected Layers) of size $1024$ with ReLU activation,
on top of which is an affine transformation of size equaling number of pixels in data ($784=28\times 28$, the size of MNIST and Fashion MNIST images).

For training details, we use $\beta=1.0$ and $x_\sigma = 0.1$ for a better quality in reconstruction.
we use Adam optimizer\citep{kingma2014adam} with learning rate $0.001$, 
$beta_1=0.9$ and $beta_2=0.999$ for optimization, and we train $100$ epochs with batch size $512$.

\paragraph{Base Classifier for MNIST and Fashion-MNIST}
The base classifier is consisting of 4 layers of FC followed by a affine transformation of size equaling to the number of labels ($10$).
We use Adam optimizer\citep{kingma2014adam} with learning rate $0.001$, $beta_1=0.9$ and $beta_2=0.999$,
and train for $100$ epochs with batch size $256$.
We use the best classifier selected from dumps at end of each epoch, based on the performance on the hold-out set.

\paragraph{Base Model and Classifier for SC09}
For SC09, we use the publicly available WaveGAN~\citep{donahue2018adversarial}\footnote{\url{https://github.com/chrisdonahue/wavegan}}
that contains pretrained GAN model for generation and classifier for classification.

\hspace{2cm}

\section{Other Approach}

\begin{figure*}[t!]
	\centering
	\includegraphics[width=0.9\textwidth,trim={0 80pt 0 0},clip]{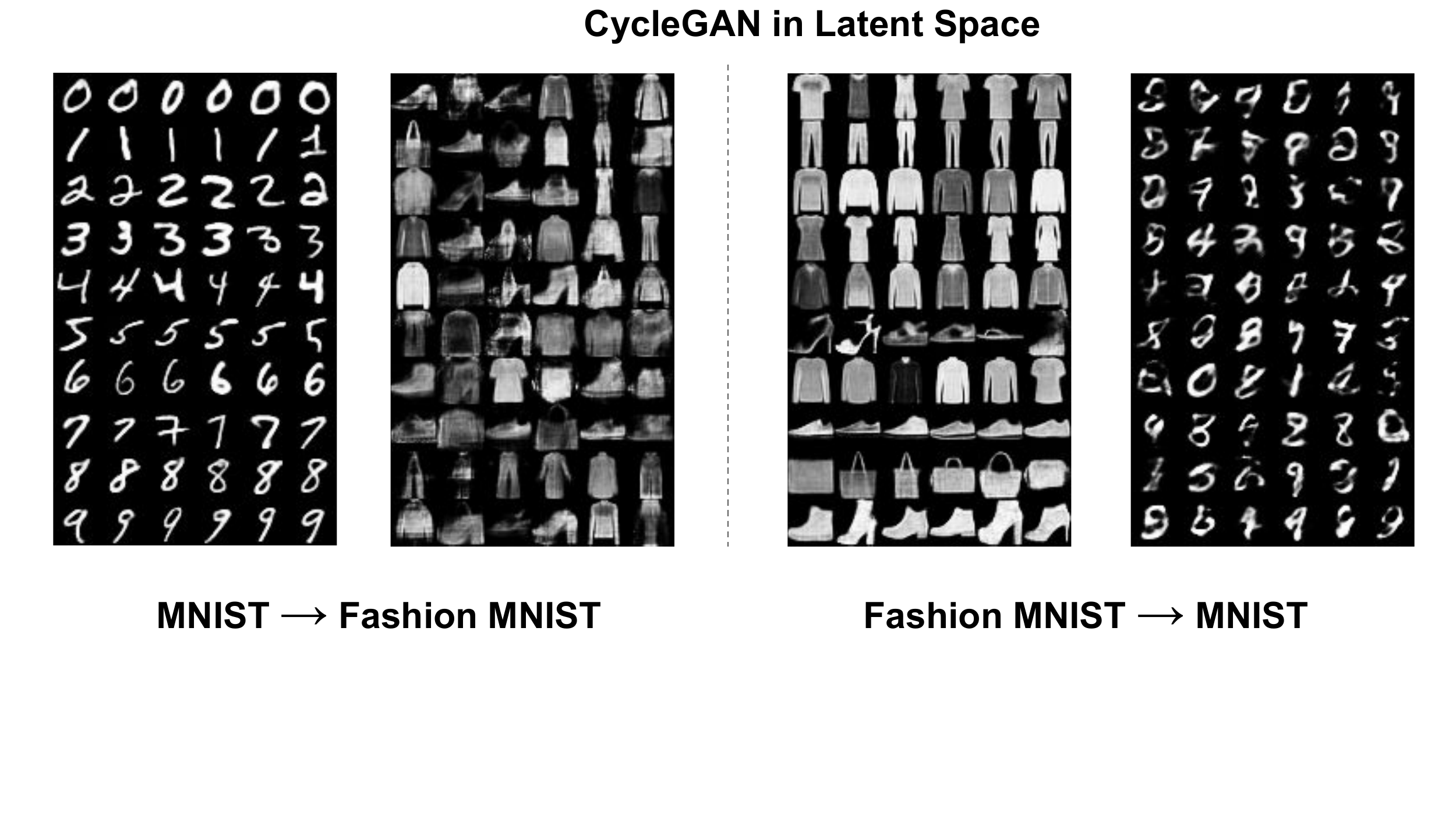}
	\caption{
		Qualitative results from applying CycleGAN~\citep{CycleGAN2017} in latent space.
		Visually, this approach suffer from less desirable overall visual quality, and most notably, failure to respect label-level supervision,
		compared to our proposed approach.
	}
	\label{fig:cyclegan-in-latent-space}
\end{figure*}

An possible alternative approach exists that applies CycleGAN in the latent space.
CycleGAN consists of two heterogeneous types of loss, the GAN and reconstruction loss,
combined using a factor $\beta$ that must be tuned.
We found that adapting CycleGAN to the latent space rather than data space leads to quite different training dynamics and therefore thoroughly tuned $\beta$.
Despite our effort, we notices that it leads to bad performance: transfer accuracy is $0.096$ for MNIST $\to$ Fashion MNIST and $0.099$ for Fashion MNIST $\to$ MNIST.
We show qualitative results from applying CycleGAN in latent space in Figure~\ref{fig:cyclegan-in-latent-space}.

\section{Supplementary Figures}

\begin{table}[h!]
	\centering
	\begin{tabular}{l l}
		\toprule
		MNIST Digits & Fashion MNIST Class \\
		\midrule
		0&	T-shirt/top \\
		1&	Trouser \\
		2&	Pullover \\
		3&	Dress \\
		4&	Coat \\
		5&	Sandal \\
		6&	Shirt \\
		7&	Sneaker \\
		8&	Bag \\
		9&	Ankle boot \\
		\bottomrule
	\end{tabular}
	\caption{MNIST Digits to Fashion MNIST class Mapping,
		made according to Labels information available at \url{https://github.com/zalandoresearch/fashion-mnist}}
	\label{table:fashion-mnist-corresponding}
\end{table}

We show the MNIST Digits to Fashion MNIST class Mapping
we used in the MNIST $\leftrightarrow$ Fashion MNIST setting detailed in the experiments
in Figure~\ref{table:fashion-mnist-corresponding}.

\end{document}